%% file: acl.tex
\pdfoutput=1
\documentclass[11pt]{article}

\usepackage[final]{acl}

\usepackage{times}
\usepackage{latexsym}

\usepackage[T1]{fontenc}
\usepackage[utf8]{inputenc}

\usepackage{microtype}

\usepackage{inconsolata}

\usepackage{newfloat}
\usepackage{listings}

\usepackage{multicol}
\usepackage{multirow}
\usepackage{tabularx}
\usepackage{booktabs}
\usepackage{enumitem}
\usepackage{amsmath}
\usepackage{nameref}
\usepackage{tcolorbox}
\usepackage{algorithm,algorithmic}
\usepackage{pbox}
\usepackage{times}
\usepackage{latexsym}
\usepackage{booktabs}
\usepackage{colortbl}
\usepackage{graphicx}
\usepackage{amssymb}
\usepackage{multirow}
\usepackage{amsmath}
\usepackage{tcolorbox}
\definecolor{backcolor}{RGB}{232, 242, 255}
\usepackage{pifont}
\newcommand{\cmark}{\ding{51}}
\newcommand{\xmark}{\ding{55}}
\newcommand{\hcmark}{{\ding{51}}\textbf{\textsuperscript{\textcolor{black}{\kern-0.55em{\bf--}}}}}

\usepackage[]{todonotes}

\title{Selection-$p$: Self-Supervised Task-Agnostic Prompt Compression for Faithfulness and Transferability}

\author{
    Tsz Ting Chung\textsuperscript{1} \quad
    Leyang Cui\textsuperscript{2} \quad
    Lemao Liu\textsuperscript{2} \quad
    Xinting Huang\textsuperscript{2} \\
    \textbf{Shuming Shi}\textsuperscript{2} \quad
    \textbf{Dit-Yan Yeung}\textsuperscript{1} \\
    \textsuperscript{1}The Hong Kong University of Science and Technology \quad
    \textsuperscript{2}Tencent AILab \\
    \texttt{\{ttchungc, nealcly.nlp, lemaoliu, shuming\}@gmail.com} \\
    \texttt{timxthuang@tencent.com} \quad
    \texttt{dyyeung@cse.ust.hk} \\
}

\begin{document}
\maketitle
\begin{abstract}

Large Language Models (LLMs) have demonstrated impressive capabilities in a wide range of natural language processing tasks when leveraging in-context learning. To mitigate the additional computational and financial costs associated with in-context learning, several prompt compression methods have been proposed to compress the in-context learning prompts.
%\neal{
Despite their success, these methods face challenges with transferability due to model-specific compression, or rely on external training data, such as GPT-4. In this paper, we investigate the ability of LLMs to develop a unified compression method that discretizes uninformative tokens, utilizing a self-supervised pre-training technique.
%}
% While recent work has explored remarkable
%architectural innovations to build longer context models, as well as 
% methods for compressing and truncating the input context, these approaches have significant drawbacks. Continuous compression into soft prompts lacks interpretability and full transferability since compression is tied to a targeted model. Meanwhile, existing discrete compression methods rely on costly external data to train compression models specifically for particular LLMs, hindering transferability\lemao{Low performance?}. To address these limitations, we propose Selection-$p$, a self-supervised pretraining approach that enables LLMs to learn how to perform discrete token-level context compression autonomously. 
%\neal{
By introducing a small number of parameters during the continual pre-training, the proposed Selection-$p$ produces %\neal{
a probability for each input token, indicating whether to preserve or discard it. Experiments show Selection-$p$ achieves state-of-the-art performance across numerous classification tasks, achieving compression rates of up to 10 times while experiencing only a marginal 0.8\% decrease in performance. Moreover, it exhibits superior transferability to different models compared to prior work. Additionally, we further analyze how Selection-$p$ helps maintain performance on in-context learning with long contexts.
% Our approach eliminates the need for costly external data or optimizing for specific LLMs, facilitating efficient and transferable context compression.
\end{abstract}
\input{sections/introduction.tex}

\input{sections/relatedwork.tex}

%%%%%%%%%%%%%%%%%%%%%%%%%%%%%%%%%%%%%%%%%%%%%%%%%%%%%%%%%%%%

%%%%%%%%%%%%%%%%%%%%%%%%%%%%%%%%%%%%%%%%%%%%%%%%%%%%%%%%%%%%

% \begin{table*}[htbp]

% {\small
% \centering
% \begin{tabular}{@{}l|ccc|ccc|ccc@{}}
% \toprule
% \multirow{2}{*}{\textbf{Methods}} & \multicolumn{3}{c|}{\textbf{MeetingBank}} & \multicolumn{6}{c}{\textbf{LongBench-Summ.}}                        \\ \cmidrule(l){2-10} 
%                                   & Summ.        & Tokens    & 1/$\tau$  & 2,000-tokens cons. & Tokens & 1/$\tau$ & 4,000-tokens cons. & Tokens & 1/$\tau$   \\ \midrule
% LLMLingua                         & 23.6        & 1,176     & 2.5x       &                    &        &         &                    &     &   \\
% LLMLingua-2                       & 30.2        & 970       & 3.0x       & 25.9               & 1,972  & 5.2x    & 27.9               & 3,923 &  2.6x \\
% Selection-$p$              & 23.1        & 1,176     & 2.5x       & 23.7               & 1,972  & 5.2x    & 25.6               & 3,923  & 2.6x  \\ \midrule
% Original Prompt                   & 26.3        & 3,003     & -          & 26.5               & 10,295 & -       & 26.5               & 10,295 & -    \\ \bottomrule
% \end{tabular}
% \caption{Evaluation with Mistral-7B on MeetingBank Summary task (Summ.) and LongBench summary task. Rouge1 is reported for the summary task.}
% \label{tb:NLG}
% }
% \end{table*}

%%%%%%%%%%%%%%%%%%%%%%%%%%%%%%%%%%%%%%%%%%%%%%%%%%%%%%%%%%%%

\input{sections/methods.tex}
\input{sections/experiment.tex}

\begin{table*}[t]
\centering
{\small
\begin{tabular}{@{}lccccccccccc@{}}
% MR   & 57.4 & 92.6 & 92.8 & +0.2 & 85.2 & 90.5 &  +5.3 & 76.1 & 74.8 & -1.3 & 71.6 & 76.6 & +5.0 & 85.1 & 82.2 &  -2.9 
\toprule
                                & Subj & RTE  & WSC  & BoolQ & MultiRC & SST-2 & WIC   & COPA & AG News & AVG  \\ \midrule
Zero-shot                       & 49.3 & 58.8 & 43.4 & 67.4  & 52.5    & 67.7  & 50.8  & 52.5 & 63.3    & 56.2 \\ \midrule
%\multicolumn{12}{c}{\textit{about 250 tokens}}                                                               \\ \midrule
Full-shot                       & 81.3 & 69.9 & 51.2 & 62.7  & 46.8    & 89.3  & 51.8  & 85.1 & 67.9    & \underline{67.3} \\
%Full-shot                      & 80.7 & 70.7 & 41.8 & 62.8  & 46.8    & 92.5  & 56.4  & 85.6 & 76.3    & \underline{68.2} \\
$\Delta_{\text{Full-shot}}$     & -0.6 & +0.8 & -9.4 &  +0.1 & $\pm$0   & +3.2  & +4.6 & +0.5 & +8.4   & +0.9 \\ \midrule
AutoCompressor                  & 56.2 & 61.5 & 44.2 & 68.3  & 52.7    & 93.0  & 51.5  & 83.6 & 76.1    & 65.2 \\
%AutoCompressor                 & 57.9 & 56.1 & 39.4 & 66.5  & 51.8    & 92.8  & 53.0  & 84.4 & 80.9    & 64.7 \\
$\Delta_{\text{AutoCompressor}}$& +1.7 & -5.4 & -4.8 & -1.8  & -0.9    & -0.2  & +1.5  & +0.8  & +4.8   & -0.5 \\ \midrule
LLMLingua*                      & 55.6 & 61.4 & 61.3 & 68.2  & 53.1    & 81.1  & 50.2  & 75.8 & 70.9    & 64.2 \\
%LLMLingua*                     & 56.7 & 60.9 & 63.0 & 69.4  & 50.3    & 69.9  & 51.6  & 76.4 & 61.5    & 62.2 \\
$\Delta_{\text{LLMLingua}}$     & +1.1 & -0.5 & +1.7 & +1.2  & -2.8    & -11.2  & +1.4 & +0.6 & -9.4    & -2.0 \\ \midrule
LLMLingua-2                     & 52.4 & 65.3 & 63.9 & 66.1  & 50.9    & 82.9  & 50.8  & 77.8 & 56.3    & 62.9 \\
%LLMLingua-2                    & 57.3 & 67.8 & 63.7 & 69.8  & 52.8    & 66.2  & 50.3  & 71.4 & 61.9    & 62.3 \\
$\Delta_{\text{LLMLingua-2}}$   & +4.9 & +2.5 & -0.2 & +3.7  & +1.9    & -16.7  & -0.5 & -6.4 & +5.6    & -0.6 \\ \midrule
\rowcolor{backcolor}
Selection-$p$            & 65.7 & 65.5 & 58.7 & 67.5  & 54.3    & 81.3  & 50.4 & 77.9 & 68.3    & \textbf{65.5} \\
%Selection-$p$           & 68.5 & 68.5 & 61.1 & 69.7  & 54.4    & 90.7  & 50.3 & 76.9 & 66.8    & \textbf{67.4} \\ 
\rowcolor{backcolor}
$\Delta_{\text{Selection-$p$}}$ & +2.8 & +3.0 & +2.4 & +2.3  & +0.1    & +9.4  & -0.1  & -1.0 & -1.5    &  +1.9\\ \bottomrule

\end{tabular}
\caption{\textbf{Performance with different number of demonstrations from about 250 tokens to about 750 tokens.} $\Delta$ refers to the performance enhancement that can be achieved by increasing the demonstration tokens size to 750.}
\label{tb:250}
}
\end{table*}

%%%%%%%%%%%%%%%%%%%%%%%%%%%%%%%%%%%%%%%%%%%%%%%%%%%%%%%%%%%%

\input{sections/analysis.tex}

%%%%%%%%%%%%%%%%%%%%%%%%%%%%%%%%%%%%%%%%%%%%%%%%%%%%%%%%%%%%

\section{Conclusion}

We introduce a simple yet effective self-supervised approach in context compression and conduct evaluation across 10 classification tasks in both few-shot and long-context settings. Our approach also demonstrated great transferability to both the open-source (i.e. LLaMA-2-13B) and black-box models (i.e. ChatGPT and Gemini), with performance surpassing the existing state-of-the-art compression models. Analysis is also conducted among different compression rates and demonstration lengths. With a 10x compression rate, our model only shows a 0.8-point drop in performance across different traditional classification tasks with a 5.3x speedup. Through experiments in long-context settings, our work also presents the possibility of addressing the in-context learning issue of the recent long-context models. Both efficiency enhancement as well as performance preservation are shown in our model.

\paragraph{Acknowledgement} This research has been made possible by the Hong Kong PhD Fellowship provided to Tsz Ting Chung and the Research Impact Fund project R6003-21 provided by the Research Grants Council of Hong Kong to Dit-Yan Yeung.
% \begin{figure*}[htbp]
%     \centering
%     \includegraphics[width=\linewidth]{figures/longcontexticl.png}
%   \caption{From another paper "Long-context LLMs Struggle with Long In-context Learning".}
% \end{figure*}
%%%%%%%%%%%%%%%%%%%%%%%%%%%%%%%%%%%%%%%%%%%%%%%%%%%%%%%%%%%%%%%%%%%% 
% \paragraph{Breaking plateauing of ICL}
% \begin{itemize}
%     \item rte task: compressed set A: 68.6->70.4(sorted), B: 65.3->69.0(sorted) A+B:71.8
%     \item only works for case sorted > unsorted
%     \item I guess:
%     \item selectionp (unsorted>sorted): formatting specific compression (more like generating a prompt --> can enhance performance but can't be used for ICL)
%     \item selectionp (sorted>unsorted): content-specific compression (benefitting ICL)
%     \item  \textcolor{red}{TODO: validate the claim}
% \end{itemize}

% \paragraph{LLMLingua-2}

% \paragraph{NLG tasks}
% summarization?
%https://aclanthology.org/2023.findings-emnlp.536.pdf

\section*{Limitations}
Under the consideration of cost, we did not perform further analysis on other LLMs apart from ChatGPT and Gemini. In addition, our model which builds up LLaMA-2-7B does not achieve better latency than models like LLMLingua-2 and AutoCompressor. Under the ICL setting, minimal time is required for compression, leading to insignificance in end-to-end inference time. While AutoCompressor offers better latency, its soft compression nature limits its applicability to other LLMs. Overall, our experiments across various tasks and settings demonstrate better performance and transferability, with the benefits outweighing the latency issue.

%However, our experiments across different tasks and settings showcase better performance and transferability.

\bibliography{custom}
\clearpage
\appendix

\section{Examples Illustration of Compression Results}
\label{sec:appendix}
%Examples of illustration for compression with Selection-$p$ in traditional classification tasks are presented in Figure \ref{fig:trad-10x} while illustration for BANKING77 is presented in Figure \ref{fig:b77} under 10x compression. On 5x compression rate, the examples of illustration for Subj task are presented in Figure \ref{fig:trad-5x-a} while for WSC task is presented in Figure \ref{fig:trad-5x-b}. In the long-context setting, we also present a 5x compression rate of BANKING77 in Figure \ref{fig:b77-5x}.

Figure \ref{fig:trad-10x} shows examples of two traditional classification tasks (i.e. Subj and WSC) with Selection-$p$ under 10x compression, displaying both the compressed results and the original demonstration sets before compression. Additionally, Figure \ref{fig:b77} illustrates an example for BANKING77 under 10x compression.

% At a 5x compression rate, Figure \ref{fig:trad-5x-a} provides examples for the Subj task, and Figure \ref{fig:trad-5x-b} provides examples for the WSC task. Figure \ref{fig:b77-5x} presents BANKING77 at a 5x compression rate in a long-context setting. 

% For a 2x compression rate, Figure \ref{fig:trad-2x-a} depicts examples for the Subj task, and Figure \ref{fig:trad-2x-b} for the WSC task. Figure \ref{fig:b77-2x} highlights the long-context setting of BANKING77 at a 2x compression rate.

\section{Prompt of Evaluation on BANKING77}
\label{sc:prompt}
In our prompt to ChatGPT, we first list out all 77 labels and then provide a list of demonstrations. The template is detailed below,\\

\hspace{-4mm}\textit{Answer in activate\_my\_card or age\_limit or apple\_pay\_or\_google\_pay or atm\_support or automatic\_top\_up or balance\_not\_updated\_after\_bank\_transfer or  ... or wrong\_amount\_of\_cash\_received or wrong\_exchange\_rate\_for\_cash\_withdrawal.}\\

\hspace{-4mm}\textit{Context: <context>}\\
\textit{Answer: <answer>}\\

\section{Training Details}
We use LLaMA-2-7B for compression (i.e. token selection). It takes roughly 50 hours on a single A100 GPU to train on 100M tokens from RedPajama.

\section{Detailed Latency Analysis}
For the token-based compression models, the time needed for compression per demonstration set on the WSC task is presented in Table \ref{tb:time}.

\begin{table}[htbp]
\centering
{\small
\begin{tabular}{@{}c|c|c@{}}
\toprule
 LLMLingua-2  & Selection-$p$ & LLMLingua  \\ \midrule
 0.15 & 0.67 & 0.81    \\
\bottomrule
\end{tabular}
\caption{Time needed for compression per demonstration set on the WSC task.} 
\label{tb:time}
}
\end{table}

Under the ICL setting, the same demonstration set is used consistently, and compression on the demonstration set only needs to be computed once for all subsequent inferences. This contributed to the short compression time in the complete end-to-end process.

Considering that the time taken for compression on the demonstration set is negligible when compared to the time required for inference (approximately at a ratio of 0.01), the end-to-end inference time for the token-selection-based compression models is roughly the same. The end-to-end inference time result for the WSC task on LLaMA-2-7B is shown in Table \ref{tb:time_w}. The table is presented in order of the inference time for clarity.

\begin{table}[htbp]
\centering
{\small
\begin{tabular}{@{}c|c|c|c@{}}
\toprule
\scriptsize AutoCompressor  & \scriptsize LLMLingua-2 & \scriptsize Selection-$p$ & \scriptsize LLMLingua \\ \midrule
22.73  & 55.6  & 55.6  & 55.6 \\
\bottomrule
\end{tabular}
\caption{End-to-end inference time on the WSC task.} 
\label{tb:time_w}
}
\end{table}
% \begin{figure}[t]

%     \begin{tcolorbox}[
%         standard jigsaw,
%         opacityback=0, 
%     ]

%     \end{tcolorbox}
%     \caption{Evaluation Prompt on BANKING77}
%     \label{fig:prompt}
% \end{figure}

\begin{figure*}[htbp]
  \begin{tcolorbox}
        \textbf{[Subj] Original Demonstrations:}\\
        input: each of the principals has a radically different way of dealing with it . \\
        type: objective\\
        
        input: well-intentioned though it may be , its soap-opera morality tales have the antiseptic , preprogrammed feel of an after-school special .\\
        type: subjective\\
        ... \\
        input: an astonishing feat for a major star let alone a 27 year old from pickum , south carolina who only two years ago was sleeping in a cardboard box in the back alleys of detroit with her mother , connie , and her uncle clutch , while playing guitar on the streets for spare change .\\
        type: objective\\
        
        input: may is a young strange girl who had a very disturbed childhood and does not still know the meaning of true friendship or love .\\
        type: objective \\\\
        \textbf{[Subj] Compressed demonstrations (10x):} \\
        input:-int it may- mor exc lord treasure planet- the ste moments- it thr en a r with he and l crowd gru a-erm own -- into imposibly ris gar ``ir wh sh de prere prede it mag m ro philosophvag rede thes v de spcer twz-passer aston fe two with cl for dist and still 
        
        \hrulefill \\\\
        \textbf{[WSC] Original Demonstrations:} \\
        Question: In the sentence "James asked Robert for a favor but he was refused.", does the pronoun 'he' refer to Robert? \\
        Answer: no\\
        
        Question: In the sentence "What about the time you cut up tulip bulbs in the hamburgers because you thought they were onions?", does the pronoun 'they' refer to tulip bulbs?\\
        Answer: yes\\
        ...\\
        Question: In the sentence "When Mr. Bond , the veterinarian, came to look at the black horse that lay groaning on the grass, he felt him all over, and shook his head; one of his legs was broken.", does the pronoun 'his' refer to the black horse?\\
        Answer: no\\
        
        Question: In the sentence "Sam took French classes from Adam , because he was eager to speak it fluently.", does the pronoun 'he' refer to Adam?\\
        Answer: no\\

        \textbf{[WSC] Compressed demonstrations (10x):} \\
        " pron:: " tul pron ' tul: he pron ' told P which P. He have pron ' P a the would only Gru une pron pronI put cfr It pron refriger man pronJohn wheng. He very im Wainws d Fol he pron veterin gro his pron the repa pron w t win gro " he pron '
    \end{tcolorbox}
    \caption{Illustration of the compression result by Selection-$p$ for Subj and WSC tasks under 10x compression rate. Compression is performed with 19 demonstrations for Subj while it is performed with 16 demonstrations for WSC with total sum of about 750 tokens respectively.}
    \label{fig:trad-10x}
\end{figure*}

\begin{figure*}[htbp]
  \begin{tcolorbox}
        \textbf{[BANKING77] Original Demonstrations:}\\
        Context: Why did using an ATM cause me to be charged an additional fee?\\
        Answer: cash\_withdrawal\_charge \\
        
        Context: I asked for a refund but its not here yet \\
        Answer: Refund\_not\_showing\_up\\
        
        Context: is there a reason i need to verify top up\\
        Answer: verify\_top\_up\\
        ...\\
        Context: There is a payment on my card that I do not recognize.  I've never seen the name on the transaction before.\\
        Answer: card\_payment\_not\_recognised     \\             
        
        Context: I happened to forget my passcode\\                                         
        Answer: passcode\_forgotten\\
        
        Context: I made a cash withdrawal and it is still listed as a pending transaction. \\
        Answer: pending\_cash\_withdrawal  \\
        
        \textbf{[BANKING77] Compressed demonstrations (10x):} \\
        c\_with asked ref Refnoting\_up\_\_recogn\_\_\_\_\_c wrong\_rece\_ transaction\_chargtw unblock activ activ\_not\_\_fe\_charg\_tim: ex sho Please revert\_\_ the card\_\_wr\_\_: I: pending\_\_:  the card\_not\_recogn: passf:\_c
    \end{tcolorbox}
    \caption{Illustration of the compression result by Selection-$p$ for BANKING77 under 10x compression rate. Compression is performed with 27 demonstrations with total sum of about 750 tokens.}
    \label{fig:b77}
\end{figure*}

\end{document}

%% file: sections/introduction.tex
\section{Introduction}

In-context learning has shown remarkable success in various natural language processing tasks~\cite{NEURIPS2020_1457c0d6}, %\neal{
such as classification task~\cite{min-etal-2022-noisy}, and mathematical reasoning task~\cite{wei2023chainofthought}, enabling Large Language Models (LLMs) to tackle complex and diverse tasks using only few-shot samples. However, in-context learning also significantly extends the length of prompts, resulting in increased computational and financial costs. Recently, a line of work has been focusing on prompt compression, which aims to compress the original prompts while minimizing information loss. They are categorized into discrete compression \cite{li-etal-2023-compressing,jiang-etal-2023-llmlingua,pan2024llmlingua2} and continuous compression \cite{NEURIPS2023_3d77c6dc,chevalier2023adapting,ge2024incontext}; the former compresses the context into discrete tokens, while the latter compresses it into a short sequence of continuous vectors.
% However, the effectiveness of in-context learning is limited by the maximum input length of Large Language Models (LLMs). 
% Recent studies attempt to address this issue by releasing long context models~\cite{liu2023lost,longlora,tworkowski2023focused} with architectural innovations in the models, as well as 
% compressing discrete~\cite{li-etal-2023-compressing,chevalier2023adapting,ge2024incontext} and continuous~\cite{NEURIPS2023_3d77c6dc,jiang-etal-2023-llmlingua,pan2024llmlingua2} contexts. In the former approaches, although longer input can be accepted into the model, performance usually declines as more context is inputted. This creates another challenge of performing compression in the context with a high compression rate while simultaneously achieving fair performance with in-context learning.

%\neal{
Observing redundant and repetitive content in a given input, discrete compression methods aim to eliminate less informative context without significantly compromising the model's performance. For example, LLMLingua~\cite{jiang-etal-2023-llmlingua} proposes to perform iterative token truncation based on the content perplexity, requiring multi-round decoding~\cite{jiang-etal-2023-llmlingua}.  LLMLingua-2 \cite{pan2024llmlingua2}, distilled from GPT-4~\cite{openai2023gpt4}, addresses the potential misalignment between entropy and the compression objective, as well as the distribution gap between the perplexity of the compression model and the target model. High costs are still involved in the training data construction. Meanwhile, optimizing the distribution for specific LLMs (GPT-4) may, on the other hand, hinder the transferability of the compressed content to other LLMs. More details are discussed in Section \ref{sc:transferable}. %(\neal{cite section xxx}.

\begin{table}[t!]
{\small
\centering
\begin{tabular}{@{}l|ccc@{}}
\toprule
                        & Transferable  &  Single Run  & $\neg$External\\ \midrule
% token for further pretraining LLMLingua-2: 29M AutoCompressor: 15B Selection-$p$: 100M
AutoCompressor          & \xmark           & \cmark    &   \cmark  \\ 
LLMLingua               & \cmark           & \xmark    &   \cmark  \\
%LLMLingua*             &                  & \cmark    &   \xmark  \\
LLMLingua-2             & \cmark           & \cmark    &   \xmark   \\
%ICAE                   &                  & \xmark    &   TODO   \\ 
\rowcolor{backcolor}
Selection-$p$    & \cmark           & \cmark   &  \cmark    \\ \bottomrule
\end{tabular}
\caption{\textbf{Comparison between the proposed Selection-$p$ and existing content compression approaches.} Selection-$p$ exhibits great transferability (Transferable), does not require multi-round iterative decoding (Single Run), and does not rely on costly external resources for training ($\neg$External).}
\label{tb: intro}
}
%\vspace{-5mm}
\end{table}

%\neal{
Continuous compression \cite{bulatov2022recurrent, wingate-etal-2022-prompt} teaches pre-trained LMs the ability to compress text into a short sequence of continuous vectors. AutoCompressors \cite{chevalier2023adapting} uses an unsupervised learning objective, which motivates the model to cache crucial information within the summary vectors. Despite their success, these methods have poor generalization as they can only compress to the length specified during training. Additionally, since the continuous vector cannot be transferred between models, a separate compressor must be trained for each model.

% Some works compress context into soft prompts and allow the reuse of the compressed soft tokens. However, this approach greatly limits the transferability of the model, as both compression and inference are performed under the same model, necessitating retraining for application to other models. Additionally, interpretability is lacking due to the difficulty in comprehending the content with continuous compressed prompts.

% To improve explainability and transferability to other models, discrete compression has been explored. While training on abstractive text compression datasets~\cite{toutanova-etal-2016-dataset,koupaee2018wikihow,kim-etal-2019-abstractive} may likely generate hallucinated content~\cite{zhao-etal-2020-reducing}, existing works mainly focus on tokens truncation from the original content. LLMLingua~\cite{jiang-etal-2023-llmlingua} proposes to perform iterative token truncation based on the content perplexity, requiring multi-round decoding~\cite{jiang-etal-2023-llmlingua}. Subsequently, \citet{pan2024llmlingua2} proposed LLMLingua-2, distilled from GPT-4~\cite{openai2023gpt4}, to address the potential misalignment between entropy and the compression objective, as well as the distribution gap between the perplexity of the compression model and the target model. However, high costs are involved in the training data construction, and optimizing the distribution for specific LLMs (GPT-4) may on the other hand hinder the transferability of the compressed content to other LLMs. Our experimental findings also support this intuition.

%\neal{
This raises an interesting research question: 

\textit{``Can LLMs learn to identify less informative tokens within a given context without external annotated signals?''}

To answer this question, %}
% These challenges raise the question: 
% \textit{Can we not rely on external annotated signals while allowing LLMs to learn to compress context autonomously?}
we propose a pre-training strategy with a self-supervised signal, which enables the model to autonomously learn to predict the next token based on compressed context. 
% how to perform context compression with Selection-$p$. 
With a small additional number of parameters, a forward pass on %\neal{
the proposed selection-$p$ creates the probability vector $p$ corresponding to each input token, indicating whether to preserve or discard the token. %\neal{
During inference, we can apply the detokenized compressed tokens to any downstream LLMs with only single-turn decoding and without reliance on any costly external resources as depicted in Table \ref{tb: intro}.
%Experiments demonstrate that our approach exhibits the best performance and transferability compared to existing state-of-the-art models while requiring only single-turn decoding during inference and without the reliance on any costly external resources as illustrated in Table \ref{tb: intro}.

The main contributions of this work are fourfold:
\begin{itemize}
    \item We present Selection-$p$ which achieves only 0.8\% drops in performance under 10x compression rate across nine traditional classification tasks, surpassing the performance of the existing compression models. Under this setting, a speedup of 5.3x can be achieved during inference with in-context learning. 
    \item Selection-$p$ demonstrated great transferability, which surpasses the performance of prior work in performing hard compression for both open-source and close-source models.
    \item We further analyze how Selection-$p$ helps in in-context learning in long-context settings, presenting a potential solution to address the performance declination of long-context models in ICL. 
    %On the declination of performance existing in current long-context models in an in-context learning setting, we show the potential solution with compression models. 
    \item We connect in-domain prior works and make comparisons with these state-of-the-art compression models, providing a complete picture.
    %and experiment on BANKING77 and 
    %Experiments on BANKING77 show that Selection-$p$ outperforms the previous compression model and some of the existing long-context models.
    
    %\item We conduct comprehensive analysis and comparison of existing compression models among different compression rates and different input token lengths (i.e. size of in-content demonstrations).
\end{itemize}

%% file: sections/relatedwork.tex
\section{Related Work}
\subsection{Hard Compression}
Some studies focus on token pruning~\cite{pmlr-v119-goyal20a,kim-cho-2021-length,rao2021dynamicvit,kim2022learned,modarressi-etal-2022-adapler} and token merging~\cite{bolya2023token} but they are designed primarily for smaller models like BERT. More recently, Selective Context~\cite{li-etal-2023-compressing} is the first to propose to prune less important tokens based on information entropy. Subsequently, LLMLingua~\cite{jiang-etal-2023-llmlingua} refined the approach by integrating the selection of demonstrations and the allocation of compression budgets for various segments of the input prompt. No training is required in these models but their efficacy in downstream tasks with compression applied to in-context demonstrations remains limited. \citet{pan2024llmlingua2} extended the idea and addressed the potential misalignment between entropy and the compression objective, leveraging full bidirectional context by training on their proposed GPT-4 distilled compression dataset. Our simple yet effective approaches outperform previous studies.

\subsection{Soft Compression}
Gist~\cite{NEURIPS2023_3d77c6dc} is first proposed to compress prompt with soft tokens. Subsequently, Autocompressor~\cite{chevalier2023adapting} and ICAE~\cite{ge2024incontext} extend the idea to handle long contexts with different pretraining approaches. ICAE further conducts instruction tuning to enhance model performance. The downstream performance of these models heavily relies on the tuned compression model, with a fixed compression rate. Additionally, retraining is necessary for different versions of LLMs. Compared to these approaches, our work offers greater flexibility and transferability, while simultaneously surpassing the performance of existing compression models.

%% file: sections/methods.tex
\section{Methodology}
Under the intuition that redundant texts often exist and their removal does not hinder human understanding of the text, we assume that LLMs behave in a similar manner. 
%\neal{
To efficiently identify less informative tokens within a given context, we propose a simple pre-training objective encouraging the model to predict the same token both before and after discarding less informative tokens.
%}
% Building upon previous works that have shown LLMs' ability to successfully recover the original text from a truncated version \cite{jiang-etal-2023-llmlingua}, we adopt a token-level pruning approach. To facilitate token pruning, we introduce an additional linear layer that assigns a probability to each token, i.e. $\theta:\{\theta_{\text{LLM}},(\mathbf{W},\mathbf{b})\}$. 
%Two different training approaches are adopted. 

%%%%%%%%%%%%%%%%%%%%%%%%%%%%%%%%%%%%%%%%%%%%%%%%%%%%%%%%%%%%%%%%%%%% 
% \subsection{Selection-$p$}

\begin{figure}[t]
    \centering
    \includegraphics[width=\linewidth]{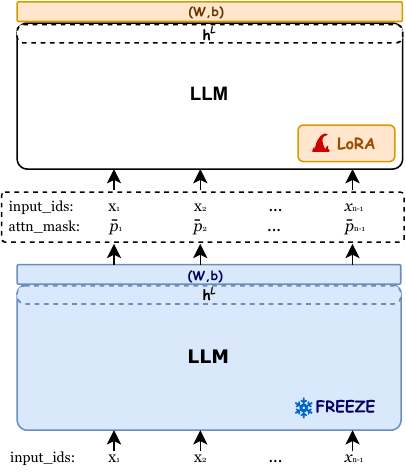}
  \caption{
    \textbf{Illustration with the training process.} Areas in orange are learnable parameters. For the input context $[x_1,x_2,\dots,x_{n-1}]$, inference without parameters update is performed first to create the attention mask $\bar{p}$. These subsequently form the model input for LoRA training and updating the parameters of the additional linear layer.}
 \label{fig:selectionp}
 \vspace{-5mm}
\end{figure}

\label{sc:selectionp}
% To learn the probability of discarding tokens, 
\subsection{Preliminary}
\paragraph{Language Model.} Given a context $[x_1,x_2,\dots,x_{n-1}]$, the objective of the language model is to predict the next token $x_n$, formed as $P(x_n|x_1,x_2,\dots,x_{n-1})$. In case of training with the causal language modeling (CLM) loss, we will have,
%\vspace{-2mm}
% \lemao{Please replace the subscript $t$ by $t$ in all the equations in this section because $t$ has been already used as }
\begin{equation*}
    \mathcal{L}_{\text{CLM}} = - \sum \log P(x_i \mid x_1,x_2, \ldots,  x_{i-1};\theta)
    %\vspace{-2mm}
\end{equation*}

\paragraph{Tokens Selection Models.}%LLMLingua.}
LLMLingua~\cite{jiang-etal-2023-llmlingua} and its variant~\cite{li-etal-2023-compressing,pan2024llmlingua2} %directly 
select tokens according to the computed %predicted 
distribution of the targeted LLM. Specifically, suppose $x_i$ is a token in the prompt, if the probability of a token $x_i$ is less than a threshold, then the token is selected to be compressed. 

\subsection{Selection-$p$}
Following the tokens selection models~\cite{li-etal-2023-compressing, jiang-etal-2023-llmlingua,pan2024llmlingua2}, Selection-$p$ includes the two steps for testing, i.e. the selection (compression) step and the inference step. 
In the {\em selection} step, %unlike LLMLingua and its variant, 
unlike LLMLingua and its variant, we instead define a selection model to select less informative tokens within the context in a discriminative way. 
In the {\em inference} step, the tokens selected via the selection model are passed to our targeted inference model.

\paragraph{Selection.}
Assume $p_i \in [0,1]$ denote a measure of the informativeness of token $x_i$. 
Theoretically, we can customize a deep neural model to instantiate $p_i$. 
In practice, we directly take a pre-trained language model and adopt the last layer of hidden representation $\mathbf{h}^L$ from a pre-trained LM to the new linear projection layer. 
Formally, suppose $\mathbf{h}^{L}=\{h_1, h_2, \cdots, h_n\}$ denotes the sequence of hidden states for all tokens $x_i$ at the last layer of a pre-trained language model. 
The selection model $\mathbf{p}=\{p_1, p_2, \cdots, p_n\}$ is defined as follows: 
\begin{equation}
   \mathbf{p} =\sigma(\mathbf{W}\mathbf{h}^{L}+{b}) 
   \label{eq:select}
\end{equation}
\noindent where $\sigma$ is the sigmoid function, $\mathbf{W}$ and $b$ are parameters of the projection matrix and bias vector, respectively. 

By using the selection model $\mathbf{p}$, it is straightforward to compress the context for inference: we directly prune the corresponding tokens according to our desired compression rate in retaining the top $k\%$ of tokens in the context. To ensure the efficiency in performing compression, a single forward pass on our model creates the token preservation probability for all input tokens for simplicity.

\paragraph{Training.}
% . This transformed representation is subsequently converted into a discarded probability value 

Our training criterion for the selection model aims to preserve the language modeling ability of the LLM while also learning to discard tokens effectively. To this end, we employ the self-supervised approach to optimize the selection model and therefore we do not need external resources to train the selection model compared with ~\citet{pan2024llmlingua2}. 
 
%($\mathbf{W}$ and $b$) of the selection model $\mathbf{p}$

To keep the training process consistent with the inference process, we first discretize the selection model $\mathbf{p}$. Let $\bar{p}_i$ denote the discretized binary model of $p_i$. In other words, $\bar{p}_i$ is 1 if it ranks the top $k\%$ of tokens with the highest $p$ values and 0 otherwise. Then we use the discretized model as a mask to define the CLM loss function as follows:
\vspace{-2mm}
\begin{equation}
    \hat{\mathcal{L}}_{\text{CLM}} = - \sum \log P(x_i \mid \bar{p}_1 x_1, \ldots, \bar{p}_{i-1} x_{i-1};\theta)
    %\vspace{-2mm}
    \label{eq:train}
\end{equation}
\noindent where $\bar{p}_ix_i$ denotes whether the token $x_i$ is masked or not depending on the value of $\bar{p}_i$, where the above language model $P$ is set as the same model as that used in the selection model in Eq.~\ref{eq:select}. 

\paragraph{Training Details for Transferability.}
One of our goals is to achieve a transferable compression method. Therefore,  the parameters that achieve the best loss on the language models in Eq.~\ref{eq:select} and Eq.~\ref{eq:train} in training may not be transferred to the targeted language model in inference since the
language models in training and inference can be different. As a result, to ensure the better transferability of the optimized selection model, we freeze the pre-trained language model in Eq.~\ref{eq:select} and employ LoRA~\cite{hu2021lora} to train a partial parameter in the language model in Eq.~\ref{eq:train}.~\footnote{In our preliminary experiments, we tried the Gumbel-softmax trick to optimize the loss in Eq.~\ref{eq:train} but we did not observe gains over the direct optimization implemented in our paper.} In summary, the CLM-based training loss is illustrated in Figure \ref{fig:selectionp}.

%%%%%%%%%%%%%%%%%%%%%%%%%%%%%%%%%%%%%%%%%%%%%%%%%%%%%%%%%%%%%%%%%%%% 
% \subsection{Pseudo Selection-$p$}
% Random sampling is performed for tokens masking, the process is repeated dozens of times in a single iteration to find the mask with the smallest language modeling loss. The resulting mask is then considered as the pseudo ``groundtruth'' $\bar{p}_{gt}$. In order to achieve the objectives of preserving language modeling capability while simultaneously truncating tokens, CLM loss in Section \ref{sc:selectionp} is applied under the pseudo ``groundtruth'' mask. At the same time, a mean squared error loss is computed to align the probability value $\textit{p}$ computed from the last hidden representation $\mathbf{h}^L$ to the ``groundtruth'' mask.

% \vspace{-3mm}
% \begin{equation*}
%     \mathcal{L} = \mathcal{L}_{\text{CLM}}(\bar{p}_{gt}) + \mathcal{L}_{\text{MSE}}(\text{\textit{p}},\bar{p}_{gt})
%     %\vspace{-2mm}
% \end{equation*}
%%%%%%%%%%%%%%%%%%%%%%%%%%%%%%%%%%%%%%%%%%%%%%%%%%%%%%%%%%%%%%%%%%%% 

%% file: sections/experiment.tex
\section{Experiment}
% Apart from the efficiency enhancement, ensuring that the compression does not harm the model capability is also important. Therefore, our model is 
% compared to other existing compression models in classification tasks, including AutoCompressor~\cite{chevalier2023adapting}, LLMLingua~\cite{jiang-etal-2023-llmlingua} %*\footnote{For LLMLingua*, the direct calling of the API does not consistently adhere to the prescribed compression ratio. To ensure a fair evaluation of both compression ratio and demonstration selection, we isolate the perplexity computation segment for comparison.}
%  and LLMLingua-2~\cite{pan2024llmlingua2} with details introduced in Section \ref{sc:baselines}.

\subsection{Setting}
We finetune a LLaMA-2-7b model~\cite{touvron2023LLaMA} on 100M tokens from RedPajama~\cite{together2023redpajama} on split segments of 1,024 tokens via LoRA~\cite{hu2022lora}. To provide a comprehensive analysis of the model capabilities, our evaluation is conducted on traditional classification tasks as well as the long-context classification task. 

For each task, we randomly sample from the training set to construct the demonstration set for In-Context Learning (ICL), which also serves as our compression target for token selection. During inference, the compression process only needs to be computed once for all subsequent inferences on the testing instances.

\paragraph{Traditional Classification Tasks.} Following \citet{chevalier2023adapting}, we evaluate and compare different compression models on nine classification tasks, including six tasks from SuperGlue~\cite{10.5555/3454287.3454581}. The predictions by LLMs are determined by iterating through all possible answer options for the instance and selecting the option with the minimum negative log-likelihood. The in-context demonstrations have been carefully selected to approximate a size of 750 tokens, and the complete demonstration is employed. This is referred to as the ``full-shot''. Since the average token length for a single demonstration varies for different tasks (e.g., a single demonstration in RTE averages about 75 tokens, resulting in a 10-shot setup under the full-shot setting), the exact number of shots differs depending on the task.
%Consequently, small variations in the size of input tokens may exist. 

To ensure a fair comparison among different compression models, a compression rate of 0.1 is used. For each task, four sets of demonstrations are selected, and the average result across these four trials is presented in Table \ref{tb:mainresult}. Additionally, the average accuracy of the nine tasks is presented for a clearer performance comparison. %This also assesses the in-context ability of the compression models. 

To achieve a compression rate of 0.1, a simpler approach is to directly retain one-tenth of the full-shot demonstration (e.g., using 1-shot instead of 10-shot for the RTE task) instead of performing selection at the token level. We also include this method as a baseline to further demonstrate the effectiveness of our approach.

\paragraph{Long Context Classification Tasks.} Recent research by~\citet{li2024longcontext} shows the failure of in-context learning tasks when applied to long-context scenarios. To investigate whether compression models can serve as a viable solution in long-context settings, we compare Selection-$p$ with long-context models, including LLaMA-2-7B-LongLora~\cite{longlora} and Long-LLaMA-code-7B~\cite{tworkowski2023focused} on the BANKING77 dataset \cite{casanueva-etal-2020-efficient}. The dataset contains 77 classes where traversing all the instances with unique labels requires approximately two thousand tokens. Evaluation is conducted at 2K, 4K, and 7K %, and 9K 
token levels, and we adopt a compression rate of 10x for Selection-$p$ and LLMLingua-2 among all levels. Since the long in-context demonstration is used, chunking is performed for every 2,048 tokens in Selection-$p$. The compressed results are concatenated together with a space token between each pair of chunks. We again follow the evaluation setting by~\citet{chevalier2023adapting} on the models' prediction, with the result presented in Table \ref{tb: longcontent}.
\subsection{Baselines}
\label{sc:baselines}
We compare the Selection-$p$ with the following state-of-the-art compression models.
\begin{itemize}
    \item \textbf{LLMLingua} \cite{jiang-etal-2023-llmlingua} employs an iterative compression algorithm to filter less informative tokens based on the token-level perplexity. To further boost the performance, \citet{jiang-etal-2023-llmlingua} also conducts a budget controller to allocate varying budgets across different demonstrations and questions. 
    % employed a Budget Controller that performs demonstration ranking and Token-level Prompt Compression to filter tokens according to the token-level perplexity. 
    We find that there is also a significant discrepancy observed between the prescribed compression rate and the actual compression rate through LLMLingua API calls. For a fair comparison, we exclusively utilize the token-level prompt compression algorithm from LLMLingua in Table~\ref{tb:mainresult}. We additionally compare LLMLingua with Selection-$p$ equipped with Budget Controller in Section \ref{sc:llmlingua}.%\neal{We additional xx}
    \item \textbf{LLMLingua-2} \cite{pan2024llmlingua2} is derived from data distillation obtained by instructing GPT-4 to perform compression. Similar to ours, the model is trained as a binary classifier on each token, determining whether each token should be preserved.
    \item \textbf{AutoCompressor} \cite{chevalier2023adapting} is constructed based on the RMT architecture~\cite{bulatov2022recurrent}. It compresses text into summary vectors that can be reused in subsequent segments. It is the only soft compression model adopted for comparison, considering that we followed the experiment setting for assessing the in-context learning ability of LLMs. 
\end{itemize}

%%%%%%%%%%%%%%%%%%%%%%%%%%%%%%%%%%%%%%%%%%%%%%%%%%%%%%%%%%%%%%%%%%%%

\begin{table*}[htbp]
\centering
{\small
\begin{tabular}{@{}lcccccccccc@{}}
% MR   & 57.4 & 92.8 & 90.5 & 92.0 & 74.8 & 76.6 & 82.2 &
\toprule
               & Subj & RTE  & WSC  & BoolQ & MultiRC & SST-2 & WIC  & COPA & AG News & AVG  \\ \midrule
Zero-shot      & 49.3 & 58.8 & 43.4 & 67.4  & 52.5    & 67.7  & 50.8 & 52.5 & 63.3    & 56.2 \\ \midrule
%\multicolumn{12}{c}{\textit{about 750 tokens}}                                                               \\ \midrule
``One-tenth''-shot & 48.6 & 65.3 & 52.6 & 68.3 & 49.2 & 84.0 & 53.6 & 83.9 & 55.5  & 62.3  \\ \midrule
Full-shot      & 80.7 & 70.7 & 41.8 & 62.8  & 46.8    & 92.5  & 56.4 & 85.6 & 76.3    & \underline{68.2} \\\midrule
AutoCompressor & 57.9 & 56.1 & 39.4 & 66.5  & 51.8    & 92.8  & 53.0 & 84.4 & 80.9    & 64.7 \\
%LLMLingua           & 67.2 & 68.8 & 55.0 & 63.6  & 48.9    & 90.5  & 52.2 & 84.7 & 73.2    & 67.1\\
LLMLingua    & 56.7 & 60.9 & 63.0 & 69.4  & 50.3    & 69.9  & 51.6 & 76.4 & 61.5    & 62.2 \\
LLMLingua-2    & 57.3 & 67.8 & 63.7 & 69.8  & 52.8    & 66.2  & 50.3 & 71.4 & 61.9    & 62.3 \\

%\textbf{Psuedo-$p$(retraining again)} 
%              & 53.6 & 69.0 & 57.0 & 67.1  &  55.2  & 83.3  & 50.8 & 76.0  & 54.2  &  62.9  \\ 
\rowcolor{backcolor}
Selection-$p$
               & 68.5 & 68.5 & 61.1 & 69.7  & 54.4    & 90.7  & 50.3 & 76.9 & 66.8    & \textbf{67.4} \\ \bottomrule
%Pseudo-\textit{p} \textcolor{orange}{retrying with a large sample no.}     & 52.6 & 67.4 & 52.1 & 66.2  & 55.0  & 67.6  & 50.9 & 79.6 & 73.8 & 53.0    & \textbf{61.8} \\ \bottomrule
%\multicolumn{12}{c}{\textit{about 250 tokens}}                                                               \\ \midrule
%Full-shot      & 81.3 & 69.9 & 51.2 & 62.7  & 46.8    & 89.3  & 51.8 & 92.6 & 85.1 & 67.9    & \underline{69.9} \\\midrule
%AutoCompressor & 56.2 & 61.5 & 44.2 & 68.3  & 52.7    & 93.0  & 51.5 & 85.2 & 83.6 & 76.1    & 67.2 \\
%LLMLingua*      & 55.6 & 61.4 & 61.3 & 68.2  & 53.1    & 81.1  & 50.2 & 76.1 & 75.8 & 70.9    & 65.4 \\
%LLMLingua-2    & 52.4 & 65.3 & 63.9 & 66.1  & 50.9    & 82.9  & 50.8 & 71.6 & 77.8 & 56.3    & 63.8 \\
%\rowcolor{backcolor}
%Selection-$p$    & 65.7 & 65.5 & 58.7 & 67.5  & 54.3    & 81.3  & 50.4 & 85.1 & 77.9 & 68.3    & \textbf{67.5} \\ \bottomrule
\end{tabular}
\caption{\textbf{Evaluation result on traditional classification tasks.} Four sets of random demonstrations were selected, with the average result being presented. The average result across different classification tasks is presented under AVG.}
\label{tb:mainresult}
}
\end{table*}

\begin{table}[t]
\centering
{\small
\begin{tabular}{@{}l|ccc@{}}
% 9K & \textbf{77.2} & 0 & 31.6 & x & 50.7
\toprule
                        & 2K   & 4K   & 7K    \\ \midrule
% LLaMA-2-7B-32K          & 30.2 & \textbf{70.4} & \textbf{75.6} \\
% LLaMA-2-7B-LongLora     & 0.0    & 0.0    & 0.0   \\
% Long-LLaMA-code-7B      & 3.0  & 19.4 & 28.0 \\ \midrule
%LLaMA-2-7B-32K          & 0.0 & 0.0 & 0.0 \\
LLaMA-2-7B-LongLora     & 0.0    & 0.0    & 0.0   \\
Long-LLaMA-code-7B      & 0.0  & 0.0 & 0.0 \\ \midrule
LLMLingua-2          & 41.2 & 31.6 & 35.2 \\
\rowcolor{backcolor}
Selection-$p$ & \textbf{46.9} & \textbf{50.9} & \textbf{51.6} \\ \bottomrule
\end{tabular}
\caption{Evaluation result on BANKING77 with increasing in-content demonstrations tokens length.} %\neal{also report the performance of autocompreesor, llmlingua, and llmlingua2!!!}}
\label{tb: longcontent}
}
\end{table}

\subsection{Evaluation Result}
%\subsubsection{}
\label{sc:icl}

\paragraph{Traditional Classification Tasks.} None of the compression models can achieve superior performance compared to the full-shot demonstration setting, which is in line with our expectations given the information loss during compression. However, certain tasks show a notable improvement when compared to both the zero-shot and full-shot approaches, e.g., all the hard compression models surpass zero-shot and full-shot by approximately 20\% in the WSC task. Among all compression models, Selection-$p$ demonstrates the highest performance in conducting ICL, with an average accuracy of 67.4\% across all 10 tasks as presented in Table \ref{tb:mainresult}. Examples of in-context demonstration before and after compression are shown in Appendix \ref{sec:appendix}. To demonstrate the effectiveness of our model, we have included one-tenth of the original demonstration set as the baseline. Our model significantly outperforms the baseline with a comparable number of tokens, therefore highlighting the effect of performing compression at the token level.

%\paragraph{Comparison to LLMLingua} 
%The result in Table \ref{tb:mainresult} also shows that ``LLMLingua*'' surpasses ``LLMLingua'' in NLU tasks, therefore we will mainly adopt ``LLMLingua*'' for comparison in the following sections.

%%%%%%%%%%%%%%%%%%%%%%%%%%%%%%%%%%%%%%%%%%%%%%%%%%%%%%%%%%%%%%%%%%%% 
% \subsection{On Natural Language Generation Tasks}
% \label{sc:NLG}
% \paragraph{Setting} We evaluate and compare different compression models for the summary tasks, including MeetingBank~\cite{hu-etal-2023-meetingbank} and LongBench~\cite{bai2023longbench} in English. We follow the setting in LLMLingua-2~\cite{pan2024llmlingua2}, keeping the token constraint and evaluating the compressed context on Mistral-7B. For MeetingBank, the original prompt tokens are limited to the thousands level. For LongBench, we adopt the long context setting and limit the original prompt tokens to the ten thousand level. Details are presented in Table \ref{tb:NLG}. Since the context provided in the two datasets is long, chunking is performed for every 2,048 tokens. The compressed results are concatenated together with a space token for each pair of chunks.

% \paragraph{Result} poor performance may due to the discreteness of the compressed tokens

%%%%%%%%%%%%%%%%%%%%%%%%%%%%%%%%%%%%%%%%%%%%%%%%%%%%%%%%%%%%%%%%%%%% 
%\subsubsection{}
\paragraph{Long Context Classification Tasks.} 
Our model outperforms LLaMA-2-7B-LongLora, Long-LLaMA-code-7B and LLMLingua-2 at all token size levels, and achieves similar results to ~\citet{li2024longcontext}'s findings on the long-context models. In addition, a growing trend with variations is observed with increasing compressed demonstrations, indicating that our model can successfully learn from additional information after compression. %However, the magnitude of growth is smaller when compared to LLaMA-2-7B-32K with increasing demonstrations. 
Examples of in-context demonstration before and after compression are presented in Appendix \ref{sec:appendix}.

% \usepackage{booktabs}

%%%%%%%%%%%%%%%%%%%%%%%%%%%%%%%%%%%%%%%%%%%%%%%%%%%%%%%%%%%%%%%%%%%% 
\subsection{Transferability}
\label{sc:transferable}
Compression is first performed on the demonstration set for ICL with Selection-$p$. Subsequently, the compressed tokens are passed to a separate downstream model (i.e. LLaMA-2-13B or the black-box models) as the compressed demonstration prompt for evaluation.

\paragraph{To LLaMA-2-13B.} We follow the same setting of evaluation across different classification tasks in Section \ref{sc:icl}. To assess the transferability of the compression models, we compress demonstrations with Selection-$p$ and input the compressed demonstration tokens into LLaMA-2-13B. In comparing different compression models, since retraining is required for soft compression methods, no results can be obtained for AutoCompressor~\cite{chevalier2023adapting}. In the case of LLMLingua, token-level perplexity is calculated with LLaMA-2-13B instead of LLaMaA-2-7B in this experiment.

Our approach outperforms all other compression models as shown in Table \ref{tb:transfer}. In addition, a small deviation is observed between the 10x compression rate and the full shot setting, demonstrating the great transferability of our models. Notably, with Selection-$p$, the tasks that outperform the full-shot setting in LLaMA-2-7b also exhibit similar patterns in LLaMA-2-13B.

\paragraph{To Black-box Models.} Taking cost into consideration, we select ChatGPT~\cite{openai2023gpt4} and Gemini~\cite{geminiteam2024gemini} for evaluation to examine its transferability to LLMs. Traditional classification tasks often have a simple nature and the potential issue of data contamination, leading to high accuracy and causing an insignificant evaluation. Therefore, we use BANKING77~\cite{casanueva-etal-2020-efficient} for evaluation. Following a similar setup as described in Section \ref{sc:icl}, we adopt a token size of 750 for examination. However, in case the compression rate is too high, ChatGPT and Gemini are more likely to deviate from the instructions and provide task-irrelevant responses. Therefore, we adopt a compression rate of 3x and use the EM metric for this experiment given their black-box nature. Note that there may be variation in the result of Gemini since the discrete compressed tokens sometimes trigger the SAFETY error. The prompt used is presented in Appendix \ref{sc:prompt}. 

Though the performance of our model still deviates from the full-shot setting, it achieved the best performance compared to the existing works as presented in Table \ref{tb:transfer}, demonstrating fair transferability even in LLMs like ChatGPT. Surprisingly, though LLMLingua-2 is distilled from GPT-4, it exhibits poor generalization compared to other compression models.

\begin{table*}[htbp]
\centering
{\small
\begin{tabular}{@{}l|cccccccccc@{}}
% MR   & 93.0 &  -   & 74.1  &  46.9  &  80.0 &
\toprule
               &  \multicolumn{10}{c}{\textbf{LLaMA-2-13B}} \\\midrule
               & Subj & RTE  & WSC  & BoolQ & MultiRC & SST-2 & WIC    & COPA & AG News & AVG \\ \midrule
%Zero-shot      & 60.4 & 68.2  & 57.7 & 73.1  & 48.1    & 84.3  & 47.2  & 80.5  & 70.5  &  65.6  \\
Full-shot      & 91.6 & 74.8 & 46.9 & 67.7  & 45.7    & 94.7  & 54.6  & 77.6 & 79.2  & \underline{70.3}\\
\midrule
%AutoCompressor &  -   &  -   &  -   &   -   &   -     &   -   &  -    &  -   &  -     &  -  & - \\
%LLMLingua      &      &      &      &       &         &       &       &      &       &     \\
LLMLingua    & 53.6 & 61.0 & 63.2 & 70.6  & 51.5    & 64.3  & 50.0  & 78.1 & 58.0  & 61.4\\
LLMLingua-2    & 48.3 & 68.7 & 41.3 & 75.8  &  52.2   & 51.3  & 49.0  & 48.2 & 71.1  &  56.2\\
% redo subj 48.3 wsc 41.3 wic 49.0 rt 46.9 copa 48.2
\rowcolor{backcolor}
Selection-$p$    
               & 69.3 & 69.5 & 65.4 & 74.7   & 50.7  & 81.1  & 50.5  & 87.5 & 63.2 & \textbf{68.0} \\ \bottomrule
\end{tabular}
\caption{\textbf{Analysis of transferability to open-source model LLaMA-2-13B.} The experiment is performed on 750 tokens in-context demonstrations with a 10x compression rate.}
\label{tb:transfer}
}
\end{table*}

\begin{table}[htbp]
\centering
{\small
\begin{tabular}{@{}l|c|c@{}}
% MR   & 93.0 &  -   & 74.1  &  46.9  &  80.0 &
\toprule
               &  \textbf{ChatGPT} & \textbf{Gemini} \\
               & GPT-3.5-Turbo     & Gemini-1.0-Pro \\\midrule
%                & \multicolumn{2}{c}{BANKING77}\\ \midrule
%Zero-shot       &  61.7          &  62.5      \\
Full-shot      & \underline{74.2} & \underline{73.3}\\
\midrule
%AutoCompressor &  -   &  -   &  -   &   -   &   -     &   -   &  -    &  -   &  -     &  -  & - \\
%LLMLingua      &      &      &      &       &         &       &       &      &       &     \\
LLMLingua     &  58.6 &  40.2\\
LLMLingua-2    &  55.7 & 51.9\\
% redo subj 48.3 wsc 41.3 wic 49.0 rt 46.9 copa 48.2
\rowcolor{backcolor}
Selection-$p$    
               & \textbf{62.9} & \textbf{58.9}\\ \bottomrule
\end{tabular}
\caption{\textbf{Analysis of transferability to blackbox models (i.e. ChatGPT and Gemini).} The experiment is performed on 750 tokens in-context demonstrations with a 3x compression rate.}
\label{tb:transfer2}
}
\end{table}

%% file: sections/analysis.tex
\section{Analysis}
%%%%%%%%%%%%%%%%%%%%%%%%%%%%%%%%%%%%%%%%%%%%%%%%%%%%%%%%%%%%%%%%%%%% 
\subsection{Flexibility}
\label{sc:flexibility}

%%%%%%%%%%%%%% COMPRESSION RATIO %%%%%%%%%%%%%%
% \paragraph{Performance with Different Compression Rates.} We examine the performance of compression models at different compression rates in classification tasks. We again follow the setting in the traditional classification tasks in Section \ref{sc:icl}, and examine across different compression rates of 2x, 5x, and 10x. %Since there are large variations in the returned token size of LLMLingua, we only examine LLMLingua* in this experiment. 
% The result is presented in Table \ref{tb:flex}.

% Selection-$p$ obtains the best performance in compression rates of 0.5 and 0.1 while LLMLingua-2 performed the best at the rate of 0.2. Overall, Selection-$p$ still demonstrated the best performance compared to other compression models. Surprisingly, the model performance does not grow with the increasing number of tokens retained. Compression rate 0f 0.5 consistently underperforms another two compression rates for all three models which potentially indicates the preservation of 50\% of tokens containing high noise information. 

%%%%%%%%%%%%%%%%%%%%%%%%%%%%%%%%%%%%%%%%%%%%%%%%%%%%%%%%%%%%%%%%%%%% 
% \subsection{P with Increasing Number of Demonstrations}
% \textcolor{red}{TODO}

%%%%%%%%%%%%%%%%%%%%%%%%%%%%%%%%%%%%%%%%%%%%%%%%%%%%%%%%%%%%%%%%%%%% 
\paragraph{Performance with Different Number of Initial Tokens.} The result in long context classification tasks in Section \ref{sc:icl} shows the effectiveness of chunk-wise compression in long context. We further analyze if compression models work well in normal few-shot settings in classification tasks. In this experiment, the in-context demonstrations are selected with an approximate size of 250 tokens. The comparison to the result with the token size of 750 in Section \ref{sc:icl} is presented in Table \ref{tb:250}.

Selection-$p$ shows the best performance under the constraint of 250 tokens when compared to other compression models. Additionally, it also follows the full-shot (i.e. 750 tokens level) trend, the average performance across all classification tasks increases along with the number of provided demonstrations. On the contrary, other compression models didn't achieve an improvement in accuracy with more demonstrations. For instance, there is a drop of 2\% recorded with an additional 500 tokens of information for AutoCompressor.

%plot: Performance with increasing \#demos vs Performance with increasing \#c\_demos

%%%%%%%%%%%%%%%%%%%%%%%%%%%%%%%%%%%%%%%%%%%%%%%%%%%%%%%%%%%%%%%%%%%% 
\subsection{Latency Analysis}
\label{sc:latency}

We analyze end-to-end latency on A100-80G GPU with the WSC task, illustrated in Table \ref{tb:latency}. Our method can achieve 5.3x speed up on 10x compressed in-context demonstration. Compared to the inference time, negligible time is required for compression on the ICL task setting, demonstrating high efficiency in adopting our models for compression. We also compared LLMLingua with the disabled content Budget Controller. It requires iterative decoding on the segmented context while Selection-$p$ only requires a single inference on all tokens and demonstrates a good performance. 
\begin{table*}[htbp]
\centering
{\small
\begin{tabular}{@{}lcccc@{}}
\toprule
                                  & 1x     & 2x     & 5x          & 10x         \\ \midrule
End-to-End {\small without compression}   & 298.6 &         &             &             \\
End-to-End {\small with Selection-$p$}  
                                  &       & 167.0 (1.8x)  & 81.6 (3.7x)    & 55.6 (5.3x)  \\ \midrule
%LLMLingua {\small / prompt}        & -     & ?    & ?        & ?        \\
LLMLingua {\small per demonstrations set}        & -     & 0.82    & 0.82        & 0.81        \\
Selection-p {\small per demonstrations set}      & -     & \textbf{0.68}    & \textbf{0.67}        & \textbf{0.67}        \\ \bottomrule
\end{tabular}
\caption{\textbf{Latency(s) comparison on WSC in 750 tokens level with about 16 demonstrations.} We present the averaged complete end-to-end inference with and without Selection-$p$ among four sets of demonstrations. Comparison is conducted with LLMLingua which also builds upon the LLaMA-2-7B backbone, with the averaged compression time of the in-context demonstrations being presented.}
\label{tb:latency}
}
\end{table*}

% \begin{table}[htbp]
% \centering
% {\small
% \begin{tabular}{@{}lcccc@{}}
% \toprule
%                                   & 1x     & 2x     & 5x          & 10x         \\ \midrule
% End2End {\tiny w/o compression}   & 298.6 &         &             &             \\
% End2End {\tiny with Selection-$p$}  
%                                   &       & 167.0   & 81.6        & 55.6      \\ 
%                                   &       & (1.8x)  &  (3.7x)     &   (5.3x) \\\midrule
% LLMLingua {\tiny / prompt}        & -     & 0.82    & 0.82        & 0.81        \\
% Selection-p {\tiny / prompt}      & -     & \textbf{0.68}    & \textbf{0.67}        & \textbf{0.67}        \\ \bottomrule
% \end{tabular}
% }
% \caption{}
% \label{tb:latency}
% \end{table}
%%%%%%%%%%%%%%%%%%%%%%%%%%%%%%%%%%%%%%%%%%%%%%%%%%%%%%%%%%%%%%%%%%%% 
\subsection{Correlation with Attention and Perplexity}

\begin{figure}[t]
    \vspace{-5mm}
    %\centering
    \includegraphics[width=0.9\linewidth]{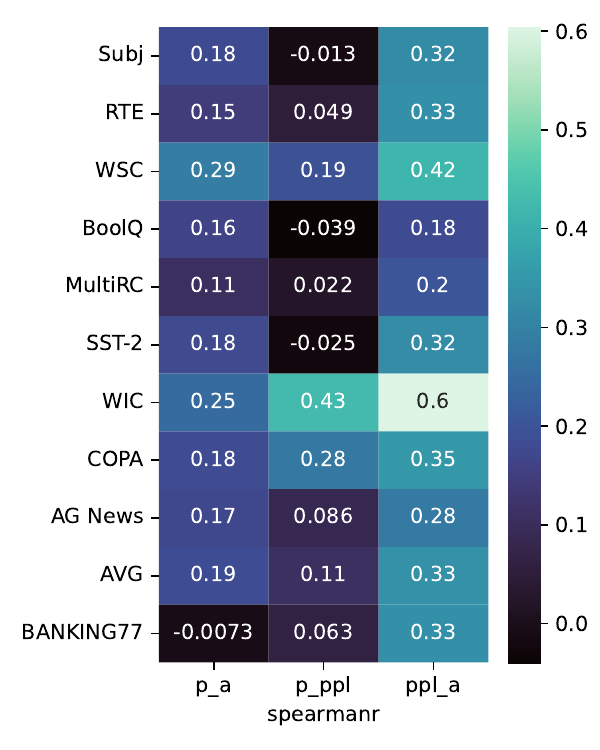}
  \caption{Spearman's Rank Correlation Coefficient (spearmanr) between \textit{p} (p) value, mean attention (a) and token-level perplexity (ppl) across different traditional classification tasks.}
 \label{fig:spearman}
 \vspace{-5mm}
\end{figure}

With the \textit{p}-value ranging between 0 and 1 for each token, we further study whether any correlations exist among \textit{p}, the mean attention value during the forward pass, and the tokens level perplexity (i.e. a core component in LLMLingua~\cite{jiang-etal-2023-llmlingua}). Since the value of \textit{p} is derived from the last hidden state of the model, we only consider the last layer mean attention of our tuned model. We employed Spearman's Rank Correlation Coefficient~\cite{spearman04} to compute the correlation between the three variables. It is calculated for different traditional classification tasks and the averaged value across tasks. The result presented in Figure \ref{fig:spearman} indicates only a weak correlation observed between the \textit{p} value and the other two variables while the correlation between the last layer mean attention and perplexity is more significant. Among all tasks, WIC demonstrates a prominently high value compared to others, this may explain the small variation in accuracy across different compression models and different experimental settings.

% \begin{table}[]
% \centering
% \begin{tabular}{@{}c|ccc@{}}
% \toprule
%                 & \textit{p}  & mean-attn.   & token-PPL \\ \midrule
% \textit{p}      & -           & 0.19  & 0.11 \\
% mean-attn.      & 0.19       & -      & 0.33  \\
% token-PPL       & 0.11       & 0.33  & -  \\ \bottomrule
% \end{tabular}
% \caption{Spearman's Rank Correlation Coefficient between \textit{p} value, mean attention (attn.) and token-level perplexity (token-PPL).}
% \label{tb:spearman}
% \end{table}

%To further analyze, we plot the top-n overlapping tokens across three variables as illustrated in Figure \ref{}. (no finding actually, so commented out)

\begin{figure*}[t]
    \centering
    \includegraphics[width=0.9\linewidth]{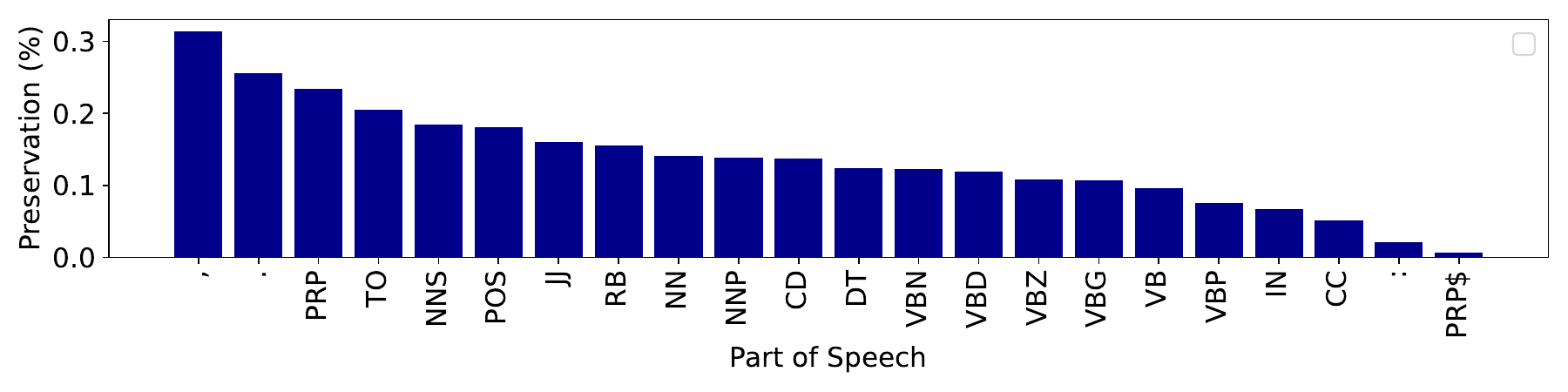}
    \caption{Analysis of the token preservation percentage with respect to different types of Part-of-Speech tags under 10x compression rate.}
    %\caption{Analysis on the distribution between frequency of Part-of-Speech tags in the compressed tokens and the original content under 10x compression rate.}
 \label{fig:pos}
 %\vspace{-5mm}
\end{figure*}

%%%%%%%%%%%%%%%%%%%%%%%%%%%%%%%%%%%%%%%%%%%%%%%%%%%%%%%%%%%%%%%%%%%% 
\subsection{Tokens Level Part-of-Speech Analysis}
To further interpret the rationale behind our compression models, we analyze what kinds of words are likely preserved by Selection-$p$. Under the discreteness of our compression result, we locate the corresponding words from the compressed tokens and obtain the Part-of-Speech (PoS) tags with an NLTK tagger. For each type of PoS tag, we compute the token preservation percentage with 
\begin{equation*}
    \frac{|\text{compressed\_token}_{\text{tag}_i}|}{|\text{total\_token}_{\text{tag}_i}|}
\end{equation*}
for each PoS tag tag$_i$. The experiment is conducted between the compressed result and the original demonstrations among the nine traditional classification tasks with four demonstration sets per task. We analyze tags with a frequency of appearance greater than 1\%.

From the result presented in Figure \ref{fig:pos}, PRP and punctuations (i.e. indicating the start of the next sentence or phrase) are more likely preserved. The potential reason for the high preservation ratio on PRP (personal pronoun) likely corresponds to the pronoun resolution task of WSC. Under the task setting, it can be useful hints for the answer derivation.

The high preservation ratio of punctuation may indicate a large redundancy in a sentence, and truncating sentence separation tokens is undesirable.
%compute the frequency of the Part-of-Speech tags of them. The frequency distribution is computed from the compressed result and the original demonstrations among the 9 traditional classification tasks across four demonstration sets per task. From the result presented in Figure \ref{fig:pos}, pronoun (PRP) and punctuations indicating the start of the next sentence or phrase are more likely preserved. It shows that large redundancy may exist in a sentence while truncating sentence separation tokens is undesirable. 
Additionally, as highlighted by \citet{wang-etal-2023-label}, formatting information (i.e. structure of the demonstrations) matters a lot in in-context learning. However, formatting tokens (i.e. ``:'') are unlikely to be preserved with Selection-$p$ compared to the original distribution in our case. In general, we also observe a higher degree of preservation of noun phrases compared to verbs.

\subsection{On Fair Comparison with LLMLingua}
\label{sc:llmlingua}
As described in Section \ref{sc:baselines}, LLMLingua conducts demonstration selection prior to compression at the token level, while other methods compress directly on the token level. Since the demonstration selection process can also be incorporated into other compression models, we only utilize the modified version of LLMLingua in the previous experiments to ensure a fair comparison.

In this section, we further analyze the performance by comparing our proposed method, equipped with the Budget Controller, with the whole LLMLingua to provide a comprehensive analysis. We follow the setting described in Section \ref{sc:icl} and select the WSC task for our experiment. To illustrate, in the original demonstration set consisting of 16 demonstrations, the LLMLingua API retains only four demonstrations. This leads to two options with Selection-$p$: (1) continuously applying the 10x compression directly to the filtered set of four demonstrations and resulting in a final compression rate of 38x, and (2) adjusting the compression rate of Selection-$p$ to achieve a final compression rate of 10x.

The result presented in Table~\ref{tb:bc} demonstrates the significant impact of the Budget Controller. Similar trends in performance for both Selection-$p$ and LLMLingua are observed (i.e., a decrease in performance on the WSC task). Notably, the performance of Selection-$p$ surpasses LLMLingua after equipping with the Budget Controller.

Furthermore, there is a disparity between the instructed compression rate and the actual compression rate in LLMLingua. The target size for compressed tokens is 75, while LLMLingua typically achieves an average compressed token size of around 192.1, which is more than 1.5 times higher than the desired rate across all classification tasks.

\begin{table}[t]
\centering
{\small
\begin{tabular}{@{}l|c|c@{}}
\toprule
                                & WSC  & rate \\ \midrule
Selection-$p$                   & 61.1 & 10x              \\
LLMLingua                      & \textbf{63.0} & 10x              \\ \midrule
Selection-$p$ \scriptsize{(+ Budget Controller)} & \textbf{47.6} & 10x              \\
Selection-$p$ \scriptsize{(+ Budget Controller)} & \textbf{57.0} & 38x              \\
LLMLingua \scriptsize{(whole)}                 & 44.7 & 10x              \\
\bottomrule
\end{tabular}
\caption{\textbf{Comparison with LLMLingua on Budget Controller.} Adopting different strategies in equipping Selection-$p$ with Budget Controller, leads to the two different compression rates (rate) of 10x and 38x. } %\lemao{Please do not use WSC dataset but another dataset such as Subj?}}
\label{tb:bc}
}
\end{table}

% \subsection{Performance Variation Along With Training}
% WSC:\\
% 50: 0.5841346153846154\\
% 150: 0.5817307692307692\\
% 250: 0.5889423076923077\\
% 350: 0.6081730769230769\\
% 450: 0.6009615384615384\\
% 550: 0.6225961538461539\\
% 650: 0.59375\\
% 750: 0.611